\documentclass[letterpaper, 10 pt, conference]{ieeeconf}  

\IEEEoverridecommandlockouts                              

\usepackage{mathptmx} 
\usepackage{amsmath} 
\usepackage{amssymb}  
\usepackage{booktabs}
 \usepackage{graphicx}
 \usepackage{float}
 \usepackage{array}
 \usepackage{tabularx}
 \usepackage[T1]{fontenc}
\usepackage[utf8]{inputenc}
 \usepackage{xcolor}
 \usepackage{url}
 \usepackage{multirow}
 \usepackage{placeins}
\usepackage{needspace}


\newif\ifreviewnotes
\reviewnotestrue

\newcommand{\paperfigure}[2]{%
  \IfFileExists{#1}{\includegraphics[width=\linewidth]{#1}}{%
    \fbox{\begin{minipage}[c][#2][c]{\dimexpr\linewidth-2\fboxsep-2\fboxrule\relax}
    \centering\small Figure file to be supplied\\[4pt]
    \texttt{\detokenize{#1}}
    \end{minipage}}}}

\title{\LARGE \bf
Motion Retargeting with Spatiotemporal Preferences for Humanoid Loco-Manipulation
}

\author{Xiongbo Shi$^{1,2,3}$, Shihang Du$^{1}$, Zongyou Yu$^{1}$, Ziyu Lin$^{1,2}$, Ao Zhou$^{1}$, Xudong Zhang$^{1}$, \\
Haodong Zhang$^{1}$, Junqiao Zhao$^{1}$, and Guang Chen$^{1,2,3,*}$%
\thanks{$^{1}$Physical Intelligence Lab, Tongji University, Shanghai, China.}%
\thanks{$^{2}$Shanghai Innovation Institute, Shanghai, China.}%
\thanks{$^{3}$RobotScale.AI.}%
\thanks{$^{*}$Corresponding author: Guang Chen (guangchen@tongji.edu.cn).}%
}

\begin{document}

\maketitle
\thispagestyle{empty}
\pagestyle{empty}

\begin{abstract}
Human motion retargeting bridges the embodiment gap between humans and humanoid robots by converting human demonstrations into reference trajectories for whole-body control and loco-manipulation. For object-interaction tasks, successful execution depends on accurately preserving body--object relations at critical stages. However, uniform per-frame iteration budgets do not account for these localized accuracy requirements, limiting the computational efficiency of retargeting. To address this limitation, we present Spatiotemporal Retargeting (STR), which uses keyframes and key body parts to allocate optimization effort. Specifically, a vision-language model selects relevant motion signals and body parts, and keyframes are identified from changes in the selected signals. Based on these selections, STR assigns additional solver iterations to keyframes and higher interaction weights to the selected parts. To further improve contact fidelity, we introduce Mechanics-Aware Contact Anchoring (MCA) to refine object-surface contact positions using an approximate object-wrench model while retaining the selected surface directions. Building on the resulting references, we develop Spatiotemporal Reinforcement Learning (STRL), which combines adaptive key-stage sampling with a contact-position reward based on target proximity for downstream control. Simulation experiments on a Unitree G1 and real-world experiments on a Unitree R1 demonstrate that STR improves geometric contact coverage and continuity at lower computational cost, while the resulting references improve downstream task completion across diverse object-interaction motions.\end{abstract}

\begin{figure*}[tp]
\centering
\paperfigure{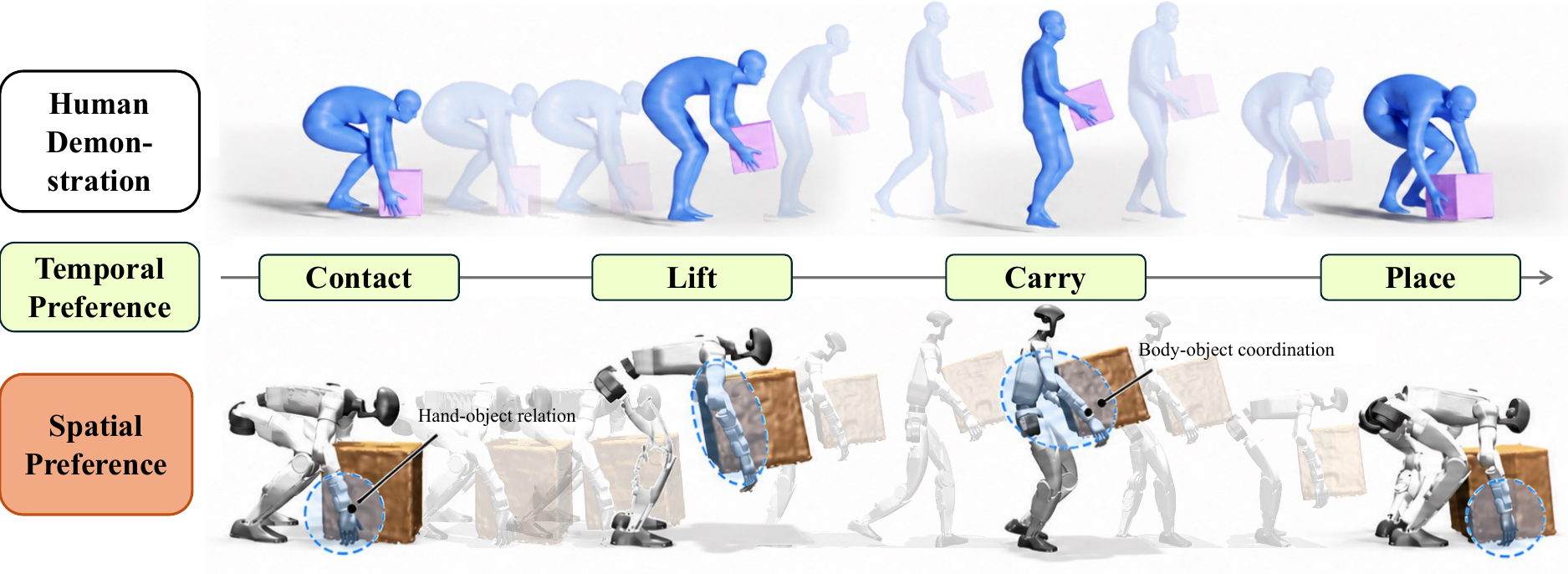}{34mm}
\caption{Keyframes and key body parts for human-to-humanoid retargeting. The human demonstration (top) and robot reference (bottom) illustrate an object-interaction motion. Highlighted poses mark keyframes, and dashed regions indicate body parts whose relation to the object is important for contact establishment and coordinated transport. STR uses these temporal and spatial priorities to allocate solver iterations and interaction weights.}
\label{fig:main}
\end{figure*}

\section{INTRODUCTION}

Human demonstrations provide a rich source of whole-body motions for learning humanoid locomotion and loco-manipulation skills~\cite{he2024h2o,he2024omnih2o,fu2025humanplus}. Parametric body models and motion collections provide reusable representations of these demonstrations~\cite{loper2015smpl,mahmood2019amass}, while interaction datasets additionally capture object motion and body--object contact~\cite{taheri2020grab,bhatnagar2022behave,li2023omomo}. However, directly transferring human motions to humanoid robots is challenging because of substantial differences in body proportions, joint configurations, and kinematic limits~\cite{pollard2002adapting,darvish2019whole}. Motion retargeting addresses this embodiment gap by converting human demonstrations into robot-compatible reference motions. Learning tracking policies from reference trajectories has enabled both simulated character control~\cite{peng2018deepmimic,luo2023phc} and humanoid whole-body control~\cite{liao2025beyondmimic}. The quality of these references directly affects how accurately and robustly the learned policy can reproduce the demonstrated behavior~\cite{araujo2025gmr}. For loco-manipulation, retargeting must additionally preserve body--object relationships that are essential for successful interaction~\cite{ho2010spatial,yang2025omniretarget,li2024chois}.

Recent retargeting methods have substantially improved the quality of such references. Video-derived references can also encode scene context for humanoid control~\cite{allshire2025videomimic}. General Motion Retargeting (GMR) reduces artifacts caused by the human--robot embodiment gap through non-uniform local scaling and inverse kinematics, and shows that retargeting artifacts can significantly affect motion-tracking robustness~\cite{araujo2025gmr}. OmniRetarget further preserves body--object--environment interactions using an interaction mesh together with kinematic and non-penetration constraints~\cite{yang2025omniretarget}. These methods improve the fidelity of the complete reference motion, but do not explicitly determine, from the demonstration, when additional retargeting accuracy is most important and which body parts should receive greater emphasis at those moments.

This distinction is particularly important for loco-manipulation. Keyframe-based motion tracking has shown that selected frames can play a distinct role in motion execution while dense references are still retained for the remaining motion~\cite{zargarbashi2025robotkeyframing,huang2025adamimic}. At the same time, the importance of different body parts changes over the course of an object-interaction motion. As illustrated in Fig.~\ref{fig:main}, accurate hand--object relations are critical when establishing contact and placing the object, whereas lifting and carrying require coordinated motion of the hands, arms, and trunk relative to the object. Errors at such keyframes or key body parts can disrupt the interaction even when the remaining reference closely follows the human demonstration. Therefore, improving the overall fidelity of a reference motion alone may not be sufficient. The retargeting process should also account for where, both in time and across the body, higher accuracy is most valuable.

In this work, we introduce \textbf{Spatiotemporal Retargeting (STR)} for efficient human-to-humanoid motion transfer. A vision-language model selects relevant motion signals and key body parts, and trajectory analysis locates the corresponding keyframes and stages. STR assigns low iteration budgets to ordinary frames, additional iterations to keyframes, and higher interaction weights to the selected body parts. This allocation preserves overall interaction fidelity with less computation. We further introduce Mechanics-Aware Contact Anchoring (MCA), which refines object-surface contact positions for the selected parts using an approximate object-wrench model. The optimized positions and fixed surface directions provide geometric tracking targets. These targets improve contact coverage and continuity while remaining within the same per-frame solver budgets.

For downstream control, we introduce \textbf{Spatiotemporal Reinforcement Learning (STRL)}. It combines failure-adaptive key-stage sampling with a contact reward that encourages each selected part to approach its target and establish contact with the object. We evaluate retargeting and policy learning with a Unitree G1 in simulation and conduct hardware experiments with a Unitree R1 and Jetson Orin NX. Our contributions are:
\begin{itemize}
\item A retargeting method that uses demonstration-derived keyframes and body-part priorities to reduce computation while retaining dense reference motions.
\item A contact-anchoring formulation that refines contact positions using object mechanics and tracks these positions and their fixed surface directions to improve geometric contact coverage and continuity.
\item A policy-learning method that combines adaptive key-stage sampling with contact-position reward shaping, together with evaluations of reference quality, computational cost, and downstream control.
\end{itemize}

\section{Related Work}
\label{sec:related_work}

\subsection{Motion Retargeting}
Motion retargeting transfers motion between bodies with different kinematics. Classical methods adapt reference motions through spacetime constraints~\cite{gleicher1998retargetting}, end-effector tracking with inverse rate control~\cite{choi1999online}, or sequential filtering with kinematic and dynamic constraints~\cite{tak2005physically}. Learning-based methods use forward kinematics and cycle consistency~\cite{villegas2018neural}, or skeleton-aware network operations~\cite{aberman2020skeleton}, to learn motion mappings across skeletons. Vision-language supervision can further preserve motion meaning by aligning embeddings of rendered source and retargeted motions~\cite{zhang2024semantics}. Beyond joint poses, interaction meshes preserve local spatial relations between a character and its surroundings~\cite{ho2010spatial}. Contact-aware retargeting preserves self-contact and ground contact while reducing self-penetration~\cite{villegas2021contact}. Cheynel et al. use sparse mesh keypoints with relevance weights that vary over space and time to describe contacts for character retargeting~\cite{cheynel2023sparse}.

For humanoid robots, early work adapts captured gestures to joint-angle and velocity limits~\cite{pollard2002adapting}. Geometric retargeting matches body-link orientations through inverse kinematics~\cite{darvish2019whole}, and interaction-mesh-based adaptation accounts for robot kinematics and dynamic balance~\cite{nakaoka2012interaction}. GMR combines non-uniform local scaling and two-stage inverse kinematics, and evaluates the effect of retargeting artifacts on motion tracking~\cite{araujo2025gmr}. VideoMimic jointly reconstructs human motion and scene geometry from video to generate references for contextual humanoid control~\cite{allshire2025videomimic}. OmniRetarget preserves body--object--environment interactions through an interaction-mesh objective with joint, velocity, non-penetration, and stance-foot constraints~\cite{yang2025omniretarget}. Our method builds on constrained interaction-preserving retargeting and uses demonstration-derived keyframes and body-part annotations to allocate solver iterations and objective weights.

\subsection{Humanoid Whole-Body Control}
Humanoid whole-body control requires coordinated motion while maintaining balance and respecting physical constraints. Motion imitation uses reference trajectories to train policies for these skills. DeepMimic combines motion-imitation and task rewards in reinforcement learning for simulated characters~\cite{peng2018deepmimic}, while AMP learns an adversarial motion prior that supplies a style reward for task execution~\cite{peng2021amp}. ASE learns reusable skill embeddings from unstructured motion data~\cite{peng2022ase}, and PHC scales imitation to large motion collections while learning recovery from falls~\cite{luo2023phc}. For humanoid robots, H2O and OmniH2O learn whole-body imitation policies for human-to-humanoid teleoperation~\cite{he2024h2o,he2024omnih2o}, and HOVER unifies multiple control interfaces through policy distillation~\cite{he2024hover}. HumanPlus uses human shadowing to collect demonstrations for learning autonomous whole-body tasks~\cite{fu2025humanplus}. ASAP learns an action correction model from real-world data and uses it to fine-tune policies in simulation~\cite{he2025asap}. BeyondMimic combines whole-body motion tracking with guided diffusion for downstream task control~\cite{liao2025beyondmimic}.

Loco-manipulation additionally requires maintaining the intended interaction with an object. HumanoidBench illustrates the difficulty of learning coordinated locomotion and manipulation in a common benchmark~\cite{sferrazza2024humanoidbench}. InterMimic learns physics-based human--object interactions from motion capture in simulation~\cite{xu2025intermimic}, while SkillMimic learns reusable basketball interaction skills by jointly imitating human and ball motion~\cite{wang2025skillmimic}. Opt2Skill trains humanoid policies to imitate dynamically feasible trajectories produced by trajectory optimization~\cite{liu2024opt2skill}. HDMI learns interactive humanoid skills from reconstructed human and object motions in video~\cite{weng2025hdmi}, while ResMimic adds object-aware residual control to a pretrained motion tracker~\cite{zhao2025resmimic}.

\subsection{Preference-Aware Motion Control}
Spatial and temporal priorities can be expressed through the body targets, contact relations, and reference frames used for motion generation and control. Spatially, ExBody emphasizes upper-body imitation while relaxing leg imitation in favor of velocity tracking~\cite{cheng2024exbody}. ExBody2 separates local keypoint tracking from velocity control~\cite{ji2026exbody2}, and HOVER uses command masks to support different combinations of body and joint targets~\cite{he2024hover}. In human--object motion synthesis, OMOMO uses hand positions as intermediate contact targets~\cite{li2023omomo}, CHOIS combines contact guidance with sparse object waypoints~\cite{li2024chois}, and InterDiff corrects predicted interactions in contact-centered coordinate frames~\cite{xu2023interdiff}. For robot object interaction, OmniRetarget preserves body--object--environment relations through an interaction mesh~\cite{yang2025omniretarget}, while ResMimic uses object-tracking and contact-tracking rewards to train an object-aware residual policy~\cite{zhao2025resmimic}. HOI-Retarget directly minimizes errors to labeled contact targets in the object frame through windowed trajectory optimization, preserving contact locations and their active intervals under kinematic limits~\cite{shin2026hoiretarget}.

Temporally, selected poses and motion segments can guide both the control objective and the training distribution. Guided time warping edits motion timing under sparse temporal constraints~\cite{hsu2007guided}. MaskedMimic controls simulated characters from partial motion specifications, including sparse keyframes and body targets~\cite{tessler2024maskedmimic}. RobotKeyframing combines sparse keyframe objectives with dense rewards for legged locomotion~\cite{zargarbashi2025robotkeyframing}. AdaMimic combines sparse global keyframe tracking with dense local tracking and learns timing and action adjustments~\cite{huang2025adamimic}. KINO uses a vision-language model to select library keyframes, aligns them to the scene, and uses motion saliency to guide keyframe sampling during policy training~\cite{chen2026kino}. BeyondMimic uses failure-based adaptive sampling in its motion-tracking stage to emphasize difficult reference segments~\cite{liao2025beyondmimic}.

Language and vision-language models offer another way to specify task-relevant relationships. Code as Policies generates executable control programs from language instructions~\cite{liang2023codeaspolicies}, and VoxPoser constructs spatial value maps for trajectory planning~\cite{huang2023voxposer}. ReKep expresses manipulation as sequences of relational keypoint constraints~\cite{huang2024rekep}, while VLMPC uses vision-language reasoning to guide candidate actions and evaluate predicted outcomes~\cite{zhao2024vlmpc}. These approaches connect high-level task descriptions to geometric or control objectives. Our method applies spatial and temporal priorities to retargeting: keyframes receive additional solver iterations, and the associated key body parts receive higher objective weights. The resulting dense references are used for policy learning with key-stage sampling.

\begin{figure*}[tp]
\centering
\paperfigure{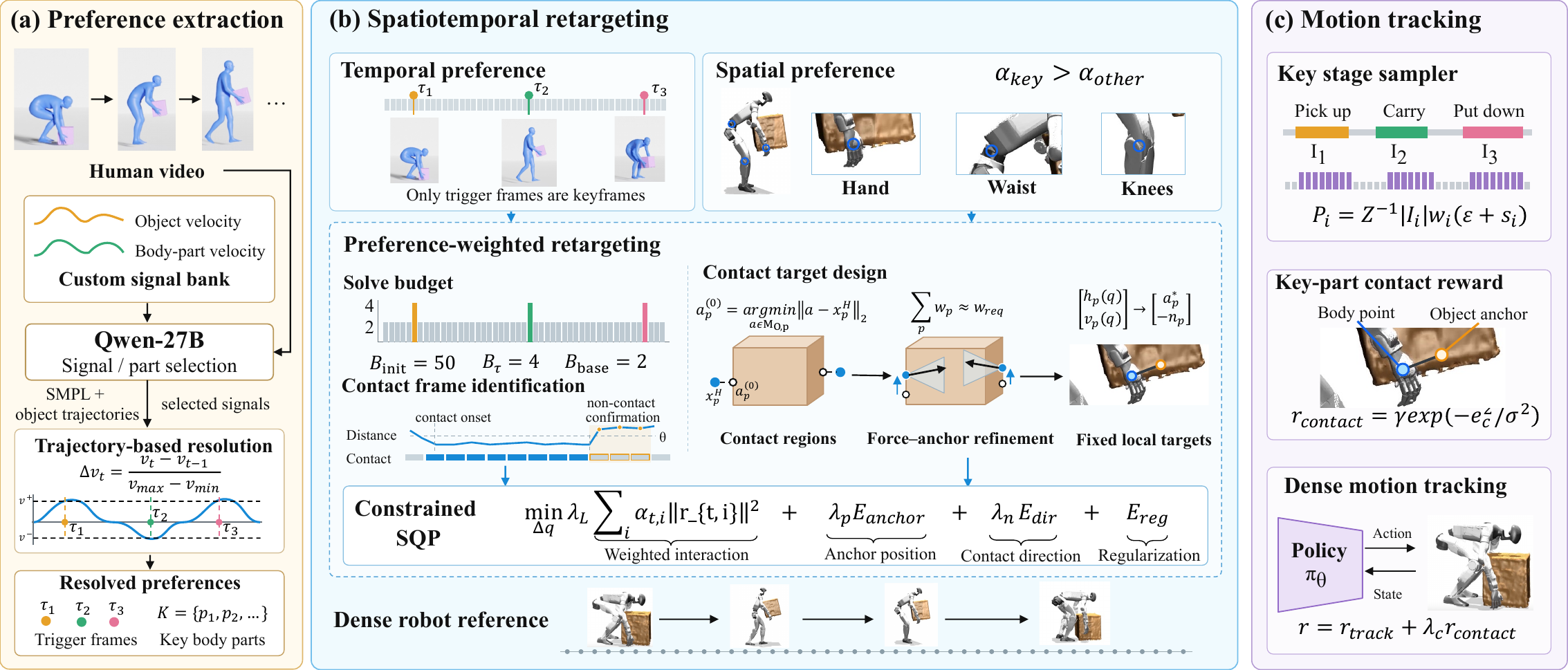}{52mm}
\caption{Overview of STR and STRL. (a) A vision-language model selects motion signals and key body parts, and trajectory analysis identifies keyframes. (b) STR allocates additional SQP iterations to keyframes and higher interaction weights to key body parts. MCA refines object contact anchors while keeping the selected surface directions fixed. Both enter the constrained optimization to produce dense robot references. (c) STRL trains a tracking policy with adaptive key-stage sampling and a contact-position reward that encourages key body parts to track their object-surface targets during the active contact intervals.}
\label{fig:method}

\end{figure*}

\section{Method}
\label{sec:method}

In this section, we introduce STR and STRL for humanoid loco-manipulation (Fig.~\ref{fig:method}). We first formulate the retargeting problem, then select motion signals and participating body parts from the demonstration video. Evaluating the selected signals on the aligned trajectories yields key stages and their trigger keyframes. STR combines these spatial and temporal priorities with contact targets designed from object geometry and motion to optimize a dense robot reference. Finally, STRL uses this reference for whole-body policy learning with key-stage sampling and contact reward shaping.

\subsection{Problem Formulation}

We retarget a human--object demonstration to a humanoid robot while preserving the body--object relations needed for task execution. The human motion and object trajectory are aligned with the demonstration video. We index the $T$ reference frames by $t=0,\ldots,T-1$ and denote the robot configuration at frame $t$ by $q_t\in\mathbb R^{n_q}$. This configuration includes the floating-base pose and joint positions. Our goal is to optimize a dense robot trajectory $\{q_t^\star\}_{t=0}^{T-1}$ under the robot's kinematic constraints, with the target object trajectory fixed throughout retargeting.

A \emph{key stage} is a segment of the demonstration associated with a task-relevant interaction, such as lifting an object. We represent a detected stage $e\in\mathrm{E}$ by its inclusive frame interval $[a_e,b_e]$, its keyframe $c_e$, and its set of key body parts $\mathrm{P}_e$. Here, $\mathrm{E}$ is the set of detected stages, and $a_e$ and $b_e$ are the start and end frames. During retargeting, $c_e$ determines where to allocate additional iterations, while $\mathrm{P}_e$ specifies which body parts receive higher relative weights within that stage.

\subsection{Keyframe and Body-Part Selection}

Applying spatiotemporal preferences requires identifying which stages and body parts are critical to the demonstrated interaction. These priorities are not explicitly specified by motion trajectories, and their timing can be difficult to determine precisely from video alone. We therefore combine video interpretation by a vision-language model (VLM) with trajectory analysis to identify key stages, their onset keyframes, and the associated key body parts.

We define a \emph{custom signal bank} for keyframe detection, providing functions that describe human and object motion. As shown in Fig.~\ref{fig:method}(a), the VLM takes the entire human video as input to identify key stages, such as \emph{Contact}, \emph{Lift}, \emph{Carry}, and \emph{Place} in Fig.~\ref{fig:main}, and select signal functions for detecting the start of each key stage. We then evaluate the selected signals on the aligned human and object trajectories through \emph{trajectory-based resolution}. Specifically, with SMPL and object trajectories as input, we compute $\Delta v_t$ between two adjacent frames for each selected signal, which will be compared with a threshold $\theta$:
\begin{equation}
    \Delta v_t = \frac{v_t-v_{t-1}}{v_{\max}-v_{\min}},
    \qquad t=1,\ldots,T-1,
    \label{eq:velocity_change}
\end{equation}
where $v_t$ is the selected velocity component at frame $t$, and $v_{\max}$ and $v_{\min}$ are its extrema over the demonstration. For the threshold $\theta$, VLM provides an initial value, which we refine during key-stage detection. The first frame of each detected key stage is its keyframe. Using the first frame as the keyframe supports sequential retargeting, where each solution initializes the next frame.

Meanwhile, the VLM selects the key body parts involved in each key stage. The selected key body parts identify where body--object relations or whole-body coordination are important during each key stage. We map these key body parts to source vertices in the interaction mesh to apply retargeting weights.

\subsection{Spatiotemporal Retargeting (STR)}
\label{sec:spatiotemporal_retargeting}

Successful object interaction requires key parts to establish the required body--object relationships at the relevant keyframes and maintain them throughout key stages. However, assigning the same iteration budget to every frame and the same weight to every body part does not reflect these spatial and temporal requirements and reduces retargeting efficiency. We therefore allocate more iterations to keyframes and assign higher weights to the body--object relationships of key parts throughout key stages, as shown in Fig.~\ref{fig:method}(b). We implement these preferences within the constrained interaction-mesh formulation~\cite{yang2025omniretarget}.

\noindent\textbf{Temporal preference.}
Our temporal preference emphasizes the onset keyframe and the subsequent frames within each key stage. Retargeting proceeds sequentially over the aligned human and object trajectories, with each optimized robot configuration initializing the next frame. We assign a larger iteration budget to the keyframe of each key stage to better preserve the body--object relationships at that frame and improve the initialization for subsequent frames. Let $\mathrm{\tau}=\{c_e:e\in\mathrm E\}$ denote the set of keyframe indices, where $\mathrm E$ is the set of detected key stages and $c_e$ is the keyframe index of stage $e$. The maximum number of sequential quadratic programming (SQP) updates at frame $t$ is
\begin{equation}
B_t=
\begin{cases}
B_{\mathrm{init}},&t=0,\\
B_{\mathrm{\tau}},&t>0,\ t\in\mathrm{\tau},\\
B_{\mathrm{base}},&\text{otherwise}.
\end{cases}
\label{eq:iteration_budget}
\end{equation}
We allocate $B_{\mathrm{init}}=50$ SQP iterations to establish the initial configuration, followed by $B_{\mathrm{\tau}}=4$ iterations at subsequent keyframes and $B_{\mathrm{base}}=2$ at ordinary frames. At each frame, optimization terminates when convergence is reached or the allocated iteration budget is exhausted.

After the keyframe, ordinary frames use the base iteration budget, but the body--object relationships of key parts must still be maintained throughout the key stage. We therefore modulate the interaction weights of key parts and their neighboring object vertices over time to retain their priority during subsequent optimization. For stage $e$, we assign higher interaction weights to key parts and their neighboring object vertices, using a temporal factor that peaks at the keyframe and decays to a positive floor:
\begin{equation}
\begin{aligned}
\bar b_e&=\min(b_e,c_e^+-1),\\
\gamma_e(t)&=\mathbf 1[c_e\le t\le\bar b_e]\,
\max\!\left\{\exp\!\left(-\frac{(t-c_e)^2}{2\sigma_w^2}\right),\rho\right\}.
\end{aligned}
\label{eq:action_weight_support}
\end{equation}
Here, $\bar b_e$ is the earlier of the stage's end frame $b_e$ and the frame immediately before the next keyframe $c_e^+$, with $c_e^+=+\infty$ when no later keyframe exists. The indicator $\mathbf 1[\cdot]$ restricts the weighting to this interval, while $\sigma_w>0$ controls the decay rate and $0<\rho\le1$ sets the minimum retained fraction of the increase. The resulting interaction weight at vertex $i$ and frame $t$ is
\begin{equation}
\begin{aligned}
\widetilde\alpha_{t,i}
 &=1+\max_{e\in\mathrm E}
 \left\{\gamma_e(t)(m_{e,i}-1)\right\},\\
\alpha_{t,i}
 &=\frac{\widetilde\alpha_{t,i}}
 {V^{-1}\sum_{j=1}^{V}\widetilde\alpha_{t,j}}.
\end{aligned}
\label{eq:vertex_weights}
\end{equation}
Here, the fixed multiplier $m_{e,i}$ is greater than one for selected body vertices and their directly adjacent object vertices and equals one elsewhere, while $\gamma_e(t)$ scales the increase $m_{e,i}-1$ over time. Parameter values are listed in Table~\ref{tab:str_parameters} (Appendix~\ref{app:str_parameters}).

At each frame, we take the maximum contribution across stages to avoid accumulating overlapping contributions and normalize the resulting weights over all $V$ graph vertices to a mean of one. These normalized weights are then applied throughout the frame's SQP updates without further adjustment, maintaining emphasis on key parts within the key stage even after the iteration budget returns to $B_{\mathrm{base}}$.

\noindent\textbf{Spatial preference.}
By assigning higher interaction weights to the representative vertices of key parts and neighboring object vertices, as defined in Eq.~\eqref{eq:vertex_weights}, we prioritize the preservation of their local body--object relationships. To promote sustained contact, we further use Mechanics-Aware Contact Anchoring (MCA) to refine object-surface contact targets for contacting key parts based on estimated contact forces, as shown in Fig.~\ref{fig:method}(b).

We initialize contact anchors from the demonstrated grasp of the contacting key parts $\mathrm P$ during their sustained contact interval $\mathrm I_c=[t_{\mathrm{on}},t_{\mathrm{off}}]$, which may span successive key stages. For the evaluated bimanual motions, where $\mathrm P$ contains both hands, we initialize each anchor at the point on a candidate object-surface region nearest to the corresponding source wrist at a representative frame within $\mathrm I_c$:
\begin{equation}
a_p^{(0)}=\underset{a\in\mathcal M_{O,p}}{\arg\min}\|a-x_p^H\|_2.
\label{eq:anchor_initialization}
\end{equation}
Here, $x_p^H$ is the source wrist position expressed in object coordinates at the selected frame, and $\mathcal M_{O,p}$ is the candidate surface region for part $p$. We jointly select opposing region pairs to minimize the total wrist-to-surface distance, with each region providing an outward direction $n_p$ that remains fixed during refinement.

Next, we estimate the required object wrench $w_t^{\mathrm{req}}$, comprising resultant force and moment, from the source object's linear and angular motion using approximate mass and inertia, with gravity included. With the anchors fixed, we compute contact forces by
\begin{equation}
\{\widehat f_{t,p}\}
=\underset{\{f_{t,p}\}\in\mathrm{F}}{\arg\min}
\sum_{t\in\mathrm I_c}
\left\|\sum_{p\in\mathrm P}w_{t,p}-w_t^{\mathrm{req}}\right\|_D^2
 +R_f.
\label{eq:contact_force_design}
\end{equation}
Here, $w_{t,p}=[f_{t,p};\,r_{t,p}\times f_{t,p}]$ stacks the world-frame contact force and its moment, with $r_{t,p}$ the moment arm from the object origin to the anchor. The set $\mathrm{F}$ enforces compressive contact and friction limits, $D$ balances force and moment errors, and $R_f$ regularizes force magnitude and temporal changes.

Finally, we fix these forces and adjust the anchors within the selected surface regions. Moving an anchor changes its moment arm, allowing the contact arrangement to reduce moment error while staying close to the initial grasp. We alternate force estimation and anchor adjustment, projecting each updated anchor onto the object surface. The refined anchors define fixed object-local position targets $a_p^\star=a_p-d_{0,p}n_p$, with a small inward offset $d_{0,p}$, and direction targets $-n_p$. These targets enter the robot optimization below. Contact interval detection, force allocation, and anchor adjustment are detailed in Appendix~\ref{app:retargeting_implementation}.

\noindent\textbf{Constrained optimization.}
The temporal and spatial preferences above determine the per-frame iteration budgets $B_t$, interaction weights $\alpha_{t,i}$, and contact targets $(a_p^\star,-n_p)$. With these quantities prescribed, we formulate retargeting as a sequence of constrained optimization problems that combine weighted interaction preservation with contact-position and contact-direction tracking.

To construct the contact objectives, forward kinematics maps each contacting part's support point and local outward direction into object coordinates, giving $h_{t,p}(q)$ and $v_{t,p}(q)$. The support point lies on the part's physical contact geometry. The desired residuals are $h_{t,p}(q)-a_p^\star$ for contact position and $v_{t,p}(q)+n_p$ for contact direction. The latter aligns the robot surface with the inward object direction, so the two contact surfaces face one another.

At SQP iteration $k$, let $q_t^{(k)}$ be the current configuration and $\Delta q$ its increment. We evaluate $h_{t,p}^{(k)}=h_{t,p}(q_t^{(k)})$ and $v_{t,p}^{(k)}=v_{t,p}(q_t^{(k)})$, with Jacobians $J_{h,t,p}^{(k)}$ and $J_{v,t,p}^{(k)}$ taken with respect to $q$. The two contact objectives are
\begin{equation}
\begin{aligned}
E_{\mathrm{anchor}}(\Delta q)
&=\sum_{p\in\mathrm P}w_p\gamma_c(t)
 \|h_{t,p}^{(k)}+J_{h,t,p}^{(k)}\Delta q-a_p^\star\|_2^2,\\
E_{\mathrm{dir}}(\Delta q)
&=\sum_{p\in\mathrm P}w_p\gamma_c(t)
 \|v_{t,p}^{(k)}+J_{v,t,p}^{(k)}\Delta q+n_p\|_2^2.
\end{aligned}
\label{eq:contact_objectives}
\end{equation}
Here, $w_p>0$ is the relative contact weight for part $p$, and $\gamma_c(t)\in[0,1]$ activates contact tracking over $\mathrm I_c$. It ramps up at $t_{\mathrm{on}}$, remains active until $t_{\mathrm{off}}$, and is zero outside the interval. Unlike the stage-dependent factor $\gamma_e(t)$, this activation sustains contact across successive key stages. Thus, $E_{\mathrm{anchor}}$ tracks the locations refined by MCA, while $E_{\mathrm{dir}}$ aligns the corresponding contact surfaces.

To maintain motion continuity while fitting these targets, we retain the temporal and joint regularization of the underlying retargeter. With trial configuration $q^+=q_t^{(k)}+\Delta q$, the regularizer is
\begin{equation}
E_{\mathrm{reg}}(\Delta q)=
\lambda_s\|q^+-q_{t-1}^\star\|_2^2
+\|Q^{1/2}q^+\|_2^2.
\label{eq:regularization}
\end{equation}
The first term penalizes deviation from the previous optimized configuration $q_{t-1}^\star$, with strength $\lambda_s$. At $t=0$, the supplied initial pose replaces $q_{t-1}^\star$. The second term regularizes selected joints toward zero through the nonnegative diagonal matrix $Q$. In the G1 implementation, only the waist yaw and roll entries are nonzero.

Combining these contact and regularization terms with the weighted interaction residuals gives
\begin{equation}
\begin{aligned}
\min_{\Delta q\in\mathrm C_t}\quad
&\lambda_L\sum_{i=1}^{V}\alpha_{t,i}\|r_{t,i}(\Delta q)\|_2^2\\
&+\lambda_p E_{\mathrm{anchor}}(\Delta q)
+\lambda_n E_{\mathrm{dir}}(\Delta q)\\
&+E_{\mathrm{reg}}(\Delta q).
\end{aligned}
\label{eq:retargeting}
\end{equation}
Here, $r_{t,i}(\Delta q)$ is the linearized difference between robot and source Laplacian coordinates at vertex $i$. Its weight $\alpha_{t,i}$ combines the temporal factor and body-part priorities in Eq.~\eqref{eq:vertex_weights}. The coefficients $\lambda_L$, $\lambda_p$, and $\lambda_n$ balance interaction preservation, anchor tracking, and direction alignment. The feasible set $\mathrm C_t$ contains linearized non-penetration and stance-foot constraints, configuration limits, and an increment bound. The object trajectory and contact targets are prescribed, and $\Delta q$ updates only the robot configuration. After each accepted update, we relinearize and continue until convergence or the budget $B_t$ in Eq.~\eqref{eq:iteration_budget} is reached. The resulting $q_t^\star$ initializes the next frame, producing the dense robot reference. Analytical Jacobians are given in Appendix~\ref{app:contact_linearization}, and solver implementation details in Appendix~\ref{app:interaction_linearization}.

\begin{table*}[tp]
\centering
\caption{Retargeting quality and computational cost on the ten-task G1 evaluation set. PECR and ECC quantify geometric contact coverage and continuity.}
\label{tab:retargeting_results}
\begingroup
\footnotesize
\setlength{\tabcolsep}{2pt}
\renewcommand{\arraystretch}{1.2}
\begin{tabular*}{\textwidth}{@{\extracolsep{\fill}}lcccccccc@{}}
\toprule
& \multicolumn{2}{c}{\multirow{2}{*}{\textbf{Penetration}}}
& \multicolumn{2}{c}{\multirow{2}{*}{\textbf{Foot Skating}}}
& \textbf{Contact} & \multicolumn{2}{c}{\textbf{Part-wise Contact}} & \textbf{Computational}\\
& \multicolumn{2}{c}{} & \multicolumn{2}{c}{}
& \textbf{Preservation} & \multicolumn{2}{c}{\textbf{(Geometric)}} & \textbf{Cost}\\
\cmidrule(lr){2-3}\cmidrule(lr){4-5}\cmidrule(lr){6-6}\cmidrule(lr){7-8}\cmidrule(lr){9-9}
\textbf{Method} & Duration (\%) $\downarrow$ & Max Depth (cm) $\downarrow$
& Duration (\%) $\downarrow$ & Max Vel. (cm/s) $\downarrow$
& Duration (\%) $\uparrow$ & PECR (\%) $\uparrow$ & ECC (\%) $\uparrow$ & SQP Iterations $\downarrow$\\
\midrule
PHC & 88.22 & 12.52 & 1.19 & 28.99 & \textbf{100.00} & 3.41 & 2.32 & --\\
GMR & 100.00 & 12.88 & 0.16 & 11.17 & \textbf{100.00} & 1.95 & 1.19 & --\\
VideoMimic & 97.52 & 12.25 & 6.17 & 64.06 & \textbf{100.00} & 0.16 & 0.12 & --\\
OmniRetarget & \textbf{0.00} & \textbf{0.00} & \textbf{0.00} & \textbf{0.00} & 99.86 & 33.65 & 24.63 & 1471.10\\
\midrule
\textbf{STR (Ours)} & 4.84 & 0.82 & \textbf{0.00} & \textbf{0.00} & 99.86 & \textbf{80.19} & \textbf{72.54} & \textbf{353.10}\\
\bottomrule
\end{tabular*}
\endgroup
\vspace{3pt}
\begin{minipage}{\textwidth}\scriptsize
\end{minipage}
\end{table*}

\subsection{Spatiotemporal Reinforcement Learning (STRL)}\label{sec:strl}

Building on the retargeted references, we introduce STRL to prioritize key and difficult stages during training and explicitly reward the selected parts for tracking their object-surface contact targets.

Inspired by~\cite{liao2025beyondmimic,yang2025omniretarget}, we use proximal policy optimization (PPO) to train a whole-body tracking policy. Specifically, the actor receives robot proprioception, reference-motion commands, and the previous action. Its actions are converted to joint-position targets for a PD controller. The critic additionally receives privileged body and object states during training. Dense body- and object-tracking rewards and the penalties for action changes, joint-limit violations, and undesired contacts are retained. The contact term augments these rewards. Training settings are provided in Appendix~\ref{app:training_configuration}.

\noindent\textbf{Allocating training starts across stages.}
After mapping the detected stage boundaries to the RL reference clock, we partition the sampleable timeline into disjoint key and ordinary stages $\{\mathrm I_j\}$. Stage $j$ includes its start frame and excludes its end frame. The end serves as a target for measuring stage completion. The final reference frame is reserved as a target and cannot be sampled as a starting frame.

Let $|\mathrm I_j|$ be the number of sampleable frames in stage $j$. Its prior weight $w_j$ is greater than one for a key stage and one for an ordinary stage. Let $f_j\in[0,1]$ estimate its failure rate. The adaptive stage distribution $\pi_j$ and the overall probability $P(j)$ of starting within stage $j$ are
\begin{equation}
\begin{aligned}
\pi_j&=\frac{|\mathrm I_j|\,w_j(f_j+\beta)}
 {\sum_{\ell}|\mathrm I_\ell|\,w_\ell(f_\ell+\beta)},\\
P(j)&=(1-\beta)\pi_j+
 \beta\frac{|\mathrm I_j|}{\sum_{\ell}|\mathrm I_\ell|}.
\end{aligned}
\label{eq:stage_sampling}
\end{equation}
Here, $\ell$ indexes all stages, and $\beta\in(0,1)$ serves both as a positive failure-rate prior and as the probability of a global uniform reset. With probability $1-\beta$, we select a stage using $\pi_j$ and then sample a frame uniformly within it. With probability $\beta$, we sample uniformly over the complete sampleable timeline. The stage-length factor makes $w_j(f_j+\beta)$ a relative priority per frame, so short stages do not receive excessive mass solely because the timeline has been partitioned. The global branch maintains coverage throughout the motion. At the sampled frame, the simulation is initialized from the reference state and the policy begins tracking the subsequent trajectory.

All $f_j$ start at zero. For an adaptively assigned stage, reaching its end records one success, while early termination before that end records one failure. We update only stages with new outcomes using $f_j\leftarrow(1-\mu)f_j+\mu\widehat f_j$, where $\widehat f_j$ is the fraction of failures among the newly resolved assignments and $\mu\in(0,1]$ is the averaging rate. Each assignment contributes at most one outcome. Reaching the stage end leaves the rollout running, and globally sampled starts do not update the stage failure estimates. This procedure initially favors key stages through $w_j$ and subsequently increases the sampling priority of difficult stages in either group.

\noindent\textbf{Key-part contact reward.}
To encourage the key body parts to establish and maintain the intended contacts, we reward proximity to the object-surface targets designed by STR, as shown in Fig.~\ref{fig:method}(c). For each contacting part $p\in\mathrm P$, we transfer the object-local position target $a_p^\star$ and contact activation to the RL reference clock, obtaining $a_{t,p}$ and $\rho_{t,p}\in[0,1]$. We pair each target with a robot-local support point $c_{t,p}$ on the training robot's contact geometry.

Let $n$ index control steps and $t=t(n)$ denote the corresponding reference frame. Using the actual simulated positions and rotations of part $p$ and the object, the world-frame support point and contact target are
\begin{equation}
\begin{aligned}
p_{n,p}&=x_{n,p}+R_{n,p}c_{t,p},\\
z_{n,p}&=x_{n,o}+R_{n,o}a_{t,p}.
\end{aligned}
\label{eq:rl_contact_points}
\end{equation}
Here, $x_{n,p},R_{n,p}$ describe the part pose, and $x_{n,o},R_{n,o}$ describe the object pose. Using the actual object pose keeps the target attached to the object during execution.

Let $e_{c,n}$ be the root-mean-square distance between the support points and their targets, weighted by contact activation, and let $\gamma_n$ be the mean activation over the active parts. The contact reward is
\begin{equation}
r_n^{\mathrm{contact}}
=\gamma_n\exp\!\left(-\frac{e_{c,n}^2}{\sigma^2}\right).
\label{eq:contact_reward}
\end{equation}
The distance scale $\sigma>0$ controls the tolerance to contact-position error, while $\gamma_n$ introduces the contact objective gradually over its prescribed interval. When no part is active, the reward is zero. The definitions of $e_{c,n}$ and $\gamma_n$, together with the contact-geometry adaptation, are provided in Appendix~\ref{app:rl_contact_details}.

We combine this contact reward with the retained tracking rewards and penalties:
\begin{equation}
r_n=r_n^{\mathrm{track}}+\lambda_c r_n^{\mathrm{contact}},
\label{eq:rl_total_reward}
\end{equation}
where $r_n^{\mathrm{track}}$ collects the baseline terms and $\lambda_c>0$ sets the relative importance of contact-position tracking.

\section{Experiments}
\label{sec:experiments}

\subsection{Experimental Setup}

\noindent\textbf{Tasks and platforms.}
We evaluate retargeting and downstream policy learning on ten diverse human--object interaction tasks from OMOMO~\cite{li2023omomo}, covering different object categories, shapes, sizes, and interaction motions. The demonstrations comprise processed SMPL motions and object trajectories. Retargeting comparisons and downstream simulation experiments use a 29-DoF Unitree G1. Hardware experiments use a Unitree R1 with policy inference on an NVIDIA Jetson Orin NX, as described in Sec.~\ref{sec:realworld}.

\noindent\textbf{Comparison methods.}
We compare STR with PHC~\cite{luo2023phc}, GMR~\cite{araujo2025gmr}, VideoMimic~\cite{allshire2025videomimic}, and OmniRetarget~\cite{yang2025omniretarget}, using the same source demonstrations, G1 model, scene geometry, and prescribed object trajectories. A four-variant ablation on the same ten tasks removes temporal and spatial preferences separately and together. Downstream comparisons use three combinations: OmniRetarget with the original RL formulation, STR with the original RL formulation, and STR with STRL. Here, original RL denotes the baseline whole-body tracking rewards and original adaptive reference sampling.

\noindent\textbf{Retargeting metrics.}
Following OmniRetarget~\cite{yang2025omniretarget}, we report \textbf{Penetration} duration and maximum depth, \textbf{Foot Skating} duration and maximum velocity, and \textbf{Contact Preservation (CP)}. CP measures whether source body--object proximity labels are retained. Our \textbf{Part-wise Effective Contact Rate (PECR)} and \textbf{Effective Contact Continuity (ECC)} assess valid surface-contact coverage and uninterrupted contact at the required body parts. Solver effort is measured by executed \textbf{SQP Iterations}. Definitions, thresholds, and formulas are provided in Appendix~\ref{app:retargeting_metrics}.

\noindent\textbf{Policy metrics.}
In simulation, we report \textbf{Success Rate (SR)}, \textbf{Object-Position RMSE}, \textbf{Object-Orientation RMSE}, \textbf{Body-Position RMSE}, and \textbf{Body-Rotation RMSE}. Success requires completing the reference trajectory without early termination, and results are computed for each task and averaged with equal weight across all ten tasks. Hardware results report successful trials out of total attempts, Body-Position RMSE, and Body-Rotation RMSE. Training settings are specified in Appendix~\ref{app:training_settings}.

\begin{figure}[t]
\centering
\paperfigure{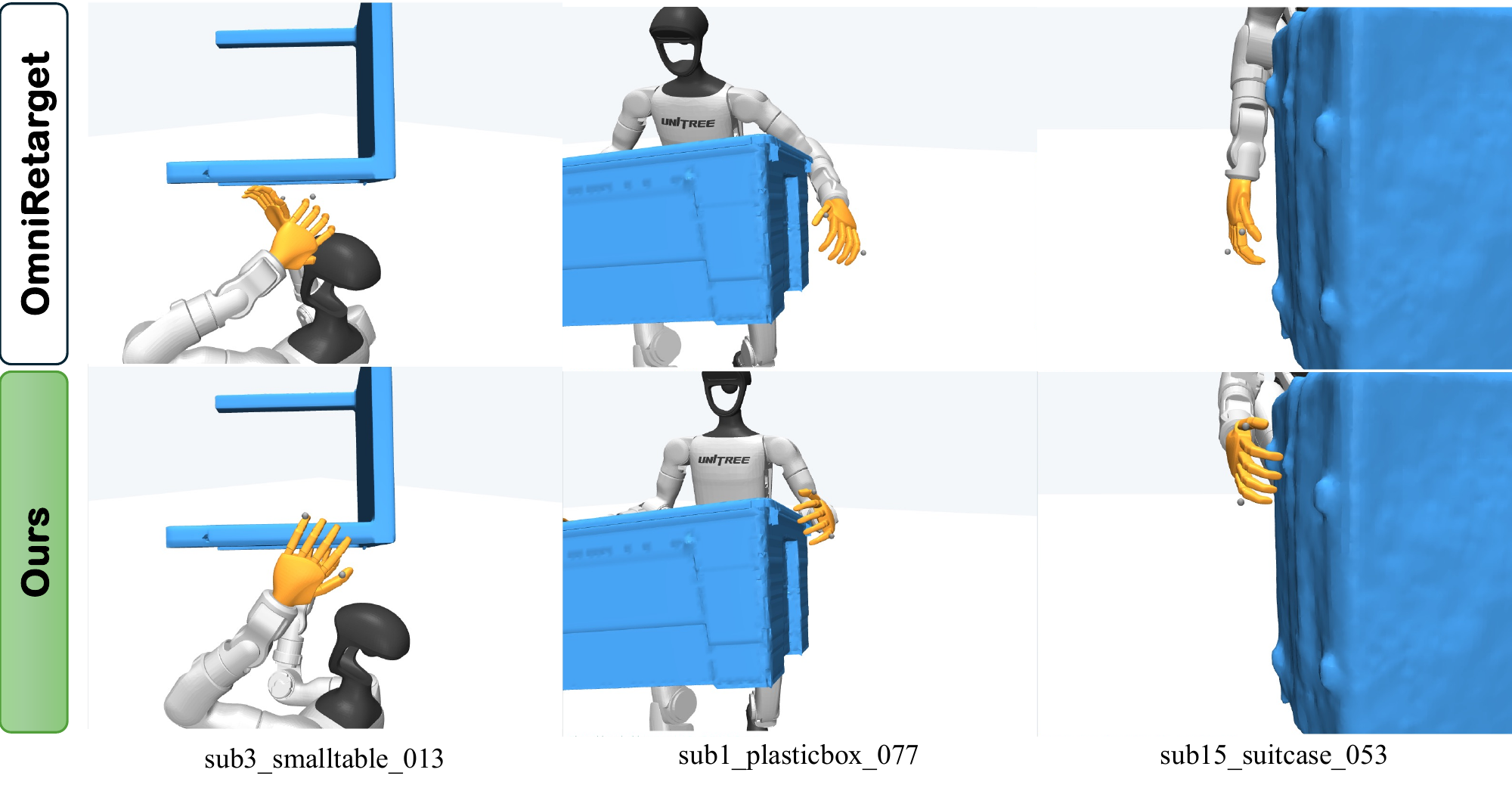}{34mm}
\caption{Qualitative comparison of retargeted human--object interactions. STR combines keyframe budgets and body-part weights with explicit contact-position and direction targets. The comparison focuses on the body--object configurations that support the demonstrated interaction.}
\label{fig:retarget_comparison}
\end{figure}

\begin{figure*}[tp]
\centering
\paperfigure{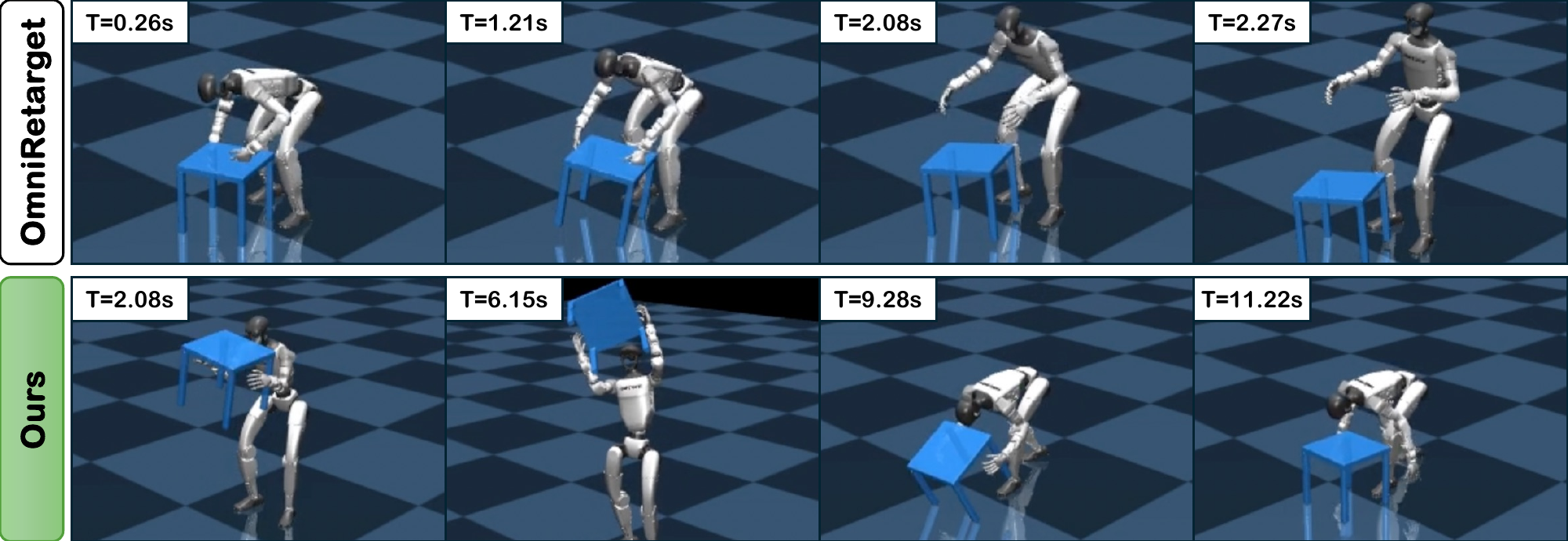}{38mm}
\caption{Effect of retargeted references on grasp execution under the original RL formulation. With OmniRetarget references, the policy fails to lift the object. With STR references, it maintains hand--object contact and completes the lift. This comparison illustrates the downstream benefit of improving reference contacts before modifying policy training.}
\label{fig:effective_contact}
\end{figure*}

\subsection{Retargeting Results}

\noindent\textbf{Contact quality.}
The quantitative results in Table~\ref{tab:retargeting_results} show that our method achieves competitive retargeting performance. In particular, it achieves the highest PECR and ECC among the compared methods, demonstrating its effectiveness in improving surface-contact coverage and continuity.

Meanwhile, the qualitative results in Fig.~\ref{fig:retarget_comparison} show that our method successfully aligns the hands with the object surface to establish contact. In contrast, OmniRetarget leaves visible hand--object gaps in the shown examples. This is because our method prioritizes key body parts and explicitly optimizes their contact positions and directions relative to the object.

\noindent\textbf{Retargeting efficiency.}
Table~\ref{tab:retargeting_results} shows that STR uses approximately 76\% fewer SQP iterations than OmniRetarget. Keyframe budgets restrict solver updates on ordinary frames while preserving additional refinement at keyframes. This reduction is maintained with MCA enabled.

\begin{table}[t]
\centering
\caption{Ablation study of spatial and temporal preferences on the same ten retargeting tasks.}
\label{tab:retargeting_ablation}
\begingroup\footnotesize
\setlength{\tabcolsep}{2pt}\renewcommand{\arraystretch}{1.15}
\begin{tabular*}{\linewidth}{@{\extracolsep{\fill}}lcccc@{}}
\toprule
\textbf{Variant} & \textbf{CP (\%)} $\uparrow$ & \textbf{PECR (\%)} $\uparrow$ & \textbf{ECC (\%)} $\uparrow$ & \textbf{SQP} $\downarrow$\\
\midrule
w/o spatial & 99.86 & 34.25 & 24.10 & 352.8\\
\midrule
w/o temporal & 99.86 & 78.32 & 68.22 & 1455.2\\
\midrule
w/o both & 99.86 & 33.65 & 24.63 & 1471.1\\
\midrule
\textbf{Full (STR)} & 99.86 & \textbf{80.19} & \textbf{72.54} & 353.1\\
\bottomrule
\end{tabular*}\endgroup
\end{table}

\noindent\textbf{Ablation study.}
Table~\ref{tab:retargeting_ablation} presents our ablation study results. Without spatial preference, i.e., with uniform interaction weights and no contact-position or contact-direction objectives, both PECR and ECC decrease substantially relative to the full method. This is because the remaining interaction-mesh objective does not explicitly align key body parts with the object surface, making contact harder to establish and maintain.

Removing temporal preference increases the iteration count yet reduces PECR and ECC, particularly ECC. This supports prioritizing keyframes and modulating interaction weights over key stages to preserve contact continuity efficiently.

Removing both spatial preference and temporal preference yields the lowest PECR and substantially reduces ECC relative to the full method. In this setting, uniform interaction weights no longer prioritize key body parts, uniform iteration budgets no longer prioritize keyframes, and the contact-position and contact-direction objectives are absent. Finally, the full method achieves the highest PECR and ECC by combining spatial preference for key body parts with temporal preference for keyframes and key stages.

\begin{figure*}[tp]
\centering
\paperfigure{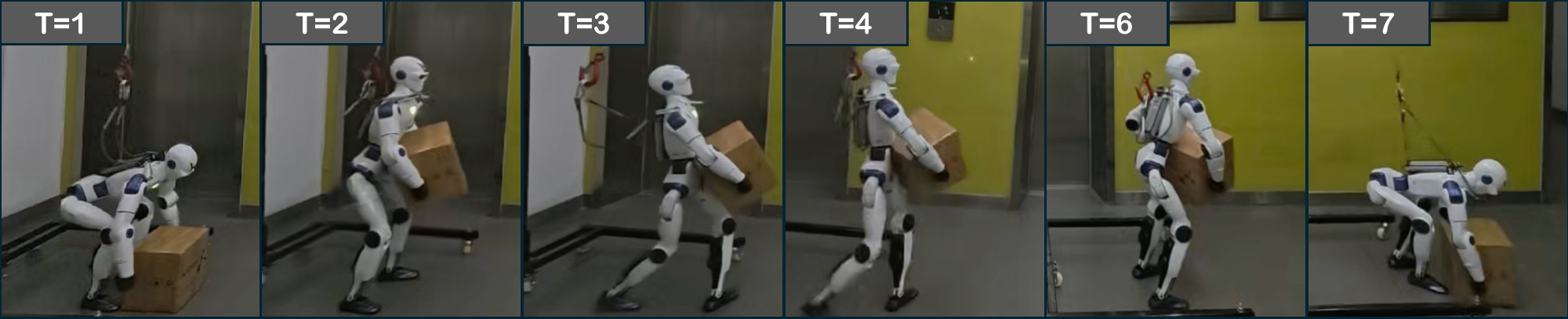}{38mm}
\caption{Real-world execution on the Unitree R1 with policy inference on an NVIDIA Jetson Orin NX. The image sequence illustrates robot--object interaction using our retargeting and policy-learning pipeline. Hardware evaluation is reported separately from the G1 simulation experiments.}
\label{fig:realworld}
\end{figure*}

\begin{table}[t]
\centering
\caption{Downstream performance on the same ten tasks in G1 simulation}
\label{tab:rl_results}
\begingroup\footnotesize
\setlength{\tabcolsep}{1.5pt}\renewcommand{\arraystretch}{1.2}
\begin{tabular*}{\linewidth}{@{\extracolsep{\fill}}lccccc@{}}
\toprule
\multirow{3}{*}{\textbf{Pipeline}} & \multirow{2}{*}{\textbf{SR}} & \multicolumn{2}{c}{\textbf{Object RMSE}} & \multicolumn{2}{c}{\textbf{Body RMSE}}\\
\cmidrule(lr){3-4}\cmidrule(lr){5-6}
& & Pos. & Rot. & Pos. & Rot.\\
& (\%) $\uparrow$ & (cm) $\downarrow$ & (${}^\circ$) $\downarrow$ & (cm) $\downarrow$ & (${}^\circ$) $\downarrow$\\
\midrule
OmniRetarget + RL & 32.97 & 21.14 & 16.81 & 7.01 & 15.72\\
\midrule
STR + RL & 81.25 & \textbf{8.80} & 11.60 & 6.85 & \textbf{15.11}\\
\midrule
STR + STRL & \textbf{86.56} & 10.00 & \textbf{7.74} & \textbf{6.82} & 16.90\\
\bottomrule
\end{tabular*}\endgroup
\vspace{3pt}
\begin{minipage}{\linewidth}\scriptsize
\end{minipage}
\end{table}

\subsection{Downstream RL Performance}

The downstream RL evaluation results in Table~\ref{tab:rl_results} show that STR improves SR and tracking performance over OmniRetarget under the original RL formulation. Adding STRL further improves SR and reduces Object-Orientation RMSE, with a slight reduction in Body-Position RMSE but higher Object-Position RMSE and Body-Rotation RMSE. These gains are consistent with better reference contacts from STR and the emphasis on key stages and contact during STRL training. Fig.~\ref{fig:effective_contact} illustrates the benefit of STR references under the original RL formulation. The policy maintains hand--object contact and completes the lift with STR references, whereas OmniRetarget references lead to contact loss and a failed lift. These results demonstrate the effectiveness of our method for humanoid object manipulation.

\subsection{Real-World Experiments}
\label{sec:realworld}

\begin{table}[t]
\centering
\caption{Box-carrying results on the Unitree R1 with policy inference on an NVIDIA Jetson Orin NX.}
\label{tab:realworld_results}
\begingroup\footnotesize
\setlength{\tabcolsep}{3pt}\renewcommand{\arraystretch}{1.2}
\begin{tabular*}{\linewidth}{@{\extracolsep{\fill}}lc@{}}
\toprule
\textbf{Metric} & \textbf{STR + STRL}\\
\midrule
Successful trials / total trials & 10 / 10\\
Body-Position RMSE (cm) $\downarrow$ & 7.16\\
Body-Rotation RMSE (${}^\circ$) $\downarrow$ & 17.30\\
\bottomrule
\end{tabular*}\endgroup
\end{table}

We evaluate the proposed framework on a box-carrying task using a Unitree R1, with policy inference running on an NVIDIA Jetson Orin NX. As shown in Fig.~\ref{fig:realworld}, the robot follows the retargeted reference motions while maintaining the intended contacts at key body parts. These contacts support box carrying as the robot coordinates its whole-body motion, providing qualitative evidence that the contact relationships established by STR can be realized during physical execution. Table~\ref{tab:realworld_results} summarizes task success and body-tracking performance. The high success rate is consistent with the qualitative results, showing that the generated reference motions and learned policies support successful interaction on the physical robot. Together, the contact behavior, task completion, and tracking results support the effectiveness of STR and STRL for real-world humanoid manipulation.

\section{Conclusion}

We presented STR, a framework for humanoid loco-manipulation that uses keyframes and key body parts to allocate optimization effort and preserve task-relevant body--object relations. Experiments demonstrate that STR improves contact coverage and continuity while reducing retargeting computation, and that the resulting references improve downstream task completion. STRL further supports policy learning through adaptive interval sampling and contact rewards.

\section{Author Contribution}

G.C. conceived the main idea.
G.C., J.Z. and H.Z. supervised the work.
G.C., X.S. and S.D. led the work.
X.S. developed the code.
X.S., S.D. and Z.L. designed the experiments.
X.S. conducted the experiments, with Z.L. contributing to the real-world experiments.
X.S. and Z.Y. wrote the manuscript.
G.C., J.Z., H.Z., S.D., A.Z. and X.Z. revised the manuscript.

\FloatBarrier
\IfFileExists{IEEEtran.bst}{\bibliographystyle{IEEEtran}}{\bibliographystyle{ieeetr}}
\bibliography{reference}

\useRomanappendicesfalse
\appendices
\flushbottom
\setlength{\jot}{2pt}
\setlength{\abovedisplayskip}{4pt plus 1pt minus 1pt}
\setlength{\belowdisplayskip}{4pt plus 1pt minus 1pt}
\setlength{\abovedisplayshortskip}{0pt plus 1pt}
\setlength{\belowdisplayshortskip}{4pt plus 1pt minus 1pt}
\setlength{\textfloatsep}{10pt plus 2pt minus 2pt}
\setlength{\intextsep}{8pt plus 1pt minus 1pt}

\section{Evaluation Metrics}
\label{app:evaluation_metrics}

\subsection{Retargeting Metrics}
\label{app:retargeting_metrics}

\noindent\textbf{Penetration and foot skating.}
Penetration is counted above 1\,cm; duration is the percentage of all reference frames with a violation. Maximum depth is the largest violating depth per task, in centimeters. Foot skating is counted when a source-designated stance foot moves horizontally by more than 0.01\,m/frame, with duration normalized by source stance frames. Maximum skating velocity converts the largest violating displacement to cm/s at 30\,Hz. Both maxima are zero when no violation occurs.

\noindent\textbf{Contact Preservation (CP).}
CP uses the baseline object-origin proximity test with a 0.28\,m radius. It reports the percentage of all reference frames with no lost source contact label, including frames without source contact. It measures proximity-label preservation; PECR and ECC assess surface-contact coverage and continuity.

\noindent\textbf{Geometric contact validity.}
For body part $k$ at frame $t$, let $A_{t,k}\in\{0,1\}$ indicate expected contact from the source demonstration. Let $\phi_{t,k}$ be the minimum signed gap between the body-part and object collision geometries, with positive values indicating separation. Let $d^{\mathrm{raw}}_{t,k}$ be the distance from the representative object-side collision endpoint to the original object surface. A geometrically valid contact satisfies
\begin{equation}
V_{t,k}=\mathbf 1[-\delta_p\leq\phi_{t,k}\leq0]\,
\mathbf 1[d^{\mathrm{raw}}_{t,k}\leq\delta_s].
\label{eq:contact_validity}
\end{equation}
We fix $\delta_p=\delta_s=10$\,mm across methods after inspecting the contact geometry. These tolerances allow shallow penetration while requiring collision-geometry contact and proximity to the original surface.

\noindent\textbf{Part-wise Effective Contact Rate (PECR).}
PECR measures valid-contact coverage over the frames in which a part is expected to contact the object:
\begin{equation}
\mathrm{PECR}_{\mathrm{geom},k}
=100\%\,\frac{\sum_t A_{t,k}V_{t,k}}{\sum_t A_{t,k}}.
\label{eq:pecr}
\end{equation}
Parts with no expected-contact frames are excluded.

\noindent\textbf{Effective Contact Continuity (ECC).}
For each maximal contiguous interval $I_{k,e}$ with $A_{t,k}=1$, let $L_{k,e}$ be the length of the longest uninterrupted run with $V_{t,k}=1$, or zero if none exists. ECC is
\begin{equation}
\mathrm{ECC}_{\mathrm{geom},k}
=100\%\,\frac{\sum_e L_{k,e}}{\sum_e|I_{k,e}|}.
\label{eq:ecc}
\end{equation}
Runs cannot cross inactive frames or separate expected-contact intervals, so fragmentation reduces ECC at a fixed valid-contact count.

For either metric $M$, we first average over the eligible parts in each task and then average equally over the ten tasks:
\begin{equation}
\overline M=\frac1N\sum_{j=1}^{N}\frac1{K_j}
\sum_{k=1}^{K_j}M_{j,k},
\label{eq:contact_average}
\end{equation}
where $N=10$ is the number of tasks and $K_j$ counts parts with expected contact in task $j$.

\noindent\textbf{SQP iterations.}
Executed SQP iterations are averaged over the ten tasks. We compare the iteration counts of OmniRetarget and STR and compute the reduction from unrounded means.

\section{Training Configuration}
\label{app:training_configuration}

\subsection{Training Settings}
\label{app:training_settings}

All comparison groups share the same G1 PPO training configuration, with 200\,Hz simulation and 50\,Hz control. Table~\ref{tab:training_parameters} lists these settings and the STRL sampling and contact-reward parameters. Actor and critic are separate MLPs with ELU activations and empirical observation normalization. The actor maps 154 inputs to 29 joint commands; the critic additionally uses privileged body and object states.

\suppressfloats[t]
\begin{table}[!htbp]
\centering
\caption{G1 PPO training settings and STRL parameters.}
\label{tab:training_parameters}
\begingroup\footnotesize
\setlength{\tabcolsep}{3pt}\renewcommand{\arraystretch}{1.06}
\begin{tabular*}{\linewidth}{@{\extracolsep{\fill}}lc@{}}
\toprule
\textbf{Parameter} & \textbf{Value}\\
\midrule
Optimizer & AdamW, zero weight decay\\
Parallel environments & 4,096\\
Rollout steps per environment & 24\\
PPO epochs / mini-batches & 5 / 4\\
Actor / critic hidden layers & 512, 256, 128\\
Activation & ELU\\
Initial actor / critic learning rate & $10^{-3}$ / $10^{-3}$\\
Learning-rate schedule / target KL & Adaptive / 0.01\\
Discount factor / GAE parameter & 0.99 / 0.95\\
PPO clipping / value-loss weight & 0.2 / 1.0\\
Entropy coefficient & 0.005\\
Gradient norm limit & 1.0\\
Initial action standard deviation & 1.0\\
Training iterations & 30,000\\
Physics / policy frequency & 200 / 50\,Hz\\
\midrule
Key / ordinary stage sampling weight & 2 / 1\\
Failure EMA rate $\mu$ & 0.001\\
Uniform reset fraction / prior $\beta$ & 0.1 / 0.1\\
Contact weight $\lambda_c$ / width $\sigma$ & 2 / 0.05\,m\\
\bottomrule
\end{tabular*}\endgroup
\end{table}

\subsection{Tracking Rewards and Regularization}
\label{app:tracking_rewards}
The baseline weights are 0.5 each for reference-anchor position and orientation, and 1.0 each for relative body position, relative body orientation, body linear velocity, body angular velocity, object position, and object orientation. The corresponding kernel widths are 0.3 for positions, 0.4 for orientations, 1.0 for linear velocity, and 3.14 for angular velocity. Action-rate, joint-limit, and undesired-contact penalties have weights $-0.1$, $-10$, and $-0.1$, respectively. Joint-limit penalties use 90\% of the position range, and undesired-contact detection uses a 1\,N threshold.

\subsection{Domain Randomization}
\label{app:domain_randomization}
During training, we randomize object mass over 0.1--2.0\,kg, object center-of-mass offsets within $\pm0.08$\,m per axis, and inertia over 50--150\% of the nominal value. Torso center-of-mass offsets are bounded by $\pm(0.025,0.05,0.075)$\,m, and joint-position bias by $\pm0.01$\,rad. Velocity perturbations are applied every 1--3\,s, with per-axis limits of 0.3\,m/s and 0.78\,rad/s. The actor observation-noise bounds are 0.05 for the reference rotation representation, 0.2\,rad/s for base angular velocity, 0.01\,rad for joint position, and 0.5\,rad/s for joint velocity. Object shape remains fixed.

\subsection{Stage-Sampling Parameters}
\label{app:stage_sampling}
Key and ordinary stages use prior weights of 2 and 1, respectively. The uniform reset fraction and failure-rate prior are both $\beta=0.1$, and the failure-rate EMA uses $\mu=0.001$.

\subsection{Contact-Reward Parameters}
\label{app:contact_reward_parameters}
The contact-position reward in Sec.~\ref{sec:strl} uses $\lambda_c=2$ and $\sigma=0.05$\,m.

\section{Retargeting Implementation}
\label{app:retargeting_implementation}

\subsection{Parameter Settings}
\label{app:str_parameters}
Table~\ref{tab:str_parameters} lists the STR settings used in the reported comparison. 

\suppressfloats[t]
\begin{table}[!htbp]
\centering
\caption{STR parameter settings for the reported G1 comparison.}
\label{tab:str_parameters}
\begingroup\footnotesize
\setlength{\tabcolsep}{3pt}\renewcommand{\arraystretch}{1.06}
\begin{tabularx}{\linewidth}{@{}>{\raggedright\arraybackslash}Xr@{}}
\toprule
\textbf{Parameter} & \textbf{Value}\\
\midrule
Stage decay width $\sigma_w$ & 2 frames\\
Minimum retained weight-increase fraction $\rho$ & 0.25\\
Selected hand / other body-vertex multiplier & 4 / 2\\
Adjacent object-vertex multiplier & 2\\
Budgets $B_{\mathrm{init}},B_{\mathrm{\tau}},B_{\mathrm{base}}$ & 50, 4, 2\\
Design mass $m_d$ & 2\,kg\\
Object-frame inertia $I_B$ & $0.04I_3$\,kg\,m$^2$\\
Friction coefficient $\mu_p$ & 0.9\\
Normal-force bounds $f^n_{p,\min},f^n_{p,\max}$ & 15, 40\,N\\
Moment-error length scale $\ell_\tau$ & 0.10\,m\\
Force-magnitude weight $\lambda_f$ & $10^{-4}$\\
Force-change weight $\lambda_{\Delta f}$ & $2\times10^{-3}$\\
Anchor-displacement scale $\ell_a$ & 0.20\,m\\
Allocation--refinement rounds & 3\\
Anchor-fit sampling stride & 2 frames\\
Contact ramp length $N_r$ & 5 frames\\
Inward target offset $d_{0,p}$ & 0.3\,mm\\
Relative contact weight $w_p$ & 1\\
Objective weights $\lambda_L,\lambda_p,\lambda_n$ & 10, 1000, 2\\
Temporal regularization weight $\lambda_s$ & 0.2\\
Waist yaw / roll diagonal entries of $Q$ & 0.2 / 0.2\\
Maximum increment norm & 0.2\\
Penetration / stance-foot XY tolerance & 1 / 1\,mm\\
\bottomrule
\end{tabularx}\endgroup
\end{table}

\subsection{Spatiotemporal Weighting and Iteration Budgets}
\label{app:weighting_budgets}
Equations~\eqref{eq:iteration_budget}--\eqref{eq:vertex_weights} use the sequence-wide constants in Table~\ref{tab:str_parameters}. At $t=0$, the initialization cap takes precedence over a keyframe cap. Selected hand vertices receive multiplier four; other selected body vertices and adjacent object vertices receive two; all others receive one. Overlapping selections retain the largest multiplier. Stage weights and contact objectives remain active at ordinary frames independently of their iteration cap. Without detected key stages, all vertex weights are one.

\subsection{Contact Interval and Activation}
\label{app:contact_interval}
Let $a_c$ and $b_c$ denote the earliest start and latest end of stages involving both hands. For each wrist, compute its unsigned distance to the object surface and apply a centered three-frame median filter with repeated endpoint values. Let $D_t$ be the larger of the two filtered wrist distances. Set $t_{\mathrm{on}}=a_c$ and define
\begin{equation}
\theta_c=
Q_{0.95}\!\left(\{D_t:t_{\mathrm{on}}\le t\le b_c\}\right)
+0.005\,\mathrm m,
\label{eq:contact_release_threshold}
\end{equation}
where $Q_{0.95}$ is the 95th percentile. After $b_c$, the first three consecutive frames with $D_t>\theta_c$ confirm release; $t_{\mathrm{off}}$ is the preceding frame, or $T-1$ if no such run exists. Over $\mathrm I_c=[t_{\mathrm{on}},t_{\mathrm{off}}]$, activation is
\begin{equation}
\begin{aligned}
s_t&=\operatorname{clip}\!\left(
\frac{t-t_{\mathrm{on}}+1}{N_r},0,1\right),\\
\gamma_c(t)&=\mathbf 1[t\in\mathrm I_c](3s_t^2-2s_t^3).
\end{aligned}
\label{eq:contact_activation}
\end{equation}
Here, $N_r$ is the ramp length and $s_t\in[0,1]$ is its normalized progress.

\subsection{Source Coordinates and Required Wrench}
\label{app:source_wrench}
Let $t_A$ be the frame of greatest source object-origin height in $\mathrm I_c$. With source wrist position $x_{t,p}^{H,W}$, object origin $z_{O,t}^H$, and object-to-world rotation $R_{O,t}^H$, the object-local wrist coordinates are
\begin{equation}
x_{t,p}^{H,O}=(R_{O,t}^H)^{\mathsf T}
             (x_{t,p}^{H,W}-z_{O,t}^H).
\label{eq:source_contact_coordinates}
\end{equation}
Superscripts $H$, $W$, and $O$ denote the demonstration, world frame, and object frame. Equation~\eqref{eq:anchor_initialization} uses $x_p^H=x_{t_A,p}^{H,O}$. We cluster object-face normals into regions and select the opposing pair minimizing the summed wrist-to-region distance, retaining their outward directions $n_p$ during refinement.

Source translation and angular velocity $\omega_t$ are smoothed with a cubic Savitzky--Golay filter using up to nine frames. Differentiation gives world translational acceleration $\ddot z_{O,t}^H$ and angular acceleration $\dot\omega_t$. With design mass $m_d$, object-frame inertia $I_B$, and gravity $g=(0,0,-9.81)^{\mathsf T}$\,m/s$^2$, the required contact wrench is
\begin{equation}
\begin{aligned}
F_t^{\mathrm{req}}&=m_d(\ddot z_{O,t}^H-g),\\
\tau_t^{\mathrm{req}}&=I_t\dot\omega_t+
                    \omega_t\times(I_t\omega_t),\\
I_t&=R_{O,t}^H I_B(R_{O,t}^H)^{\mathsf T}.
\end{aligned}
\label{eq:required_wrench}
\end{equation}
These components form $w_t^{\mathrm{req}}$ in Eq.~\eqref{eq:contact_force_design}, with moments about the object origin and world inertia $I_t$. The approximate design mass and isotropic inertia are given in Table~\ref{tab:str_parameters}; $I_3$ is the $3\times3$ identity.

\subsection{Force Allocation}
\label{app:force_allocation}
For fixed anchors, define world moment arms $r_{t,p}$, inward unit directions $\nu_{t,p}$, normal-force components $f^n_{t,p}$, and tangential-force vectors $f^\parallel_{t,p}$ as
\begin{equation}
\begin{aligned}
r_{t,p}&=R_{O,t}^H a_p,&
\nu_{t,p}&=-R_{O,t}^H n_p,\\
f^n_{t,p}&=\nu_{t,p}^{\mathsf T}f_{t,p},&
f^\parallel_{t,p}&=f_{t,p}-f^n_{t,p}\nu_{t,p}.
\end{aligned}
\label{eq:contact_force_components}
\end{equation}
The normal component is positive for compression into the object. Forces are allocated jointly over the contact interval:
\begin{equation}
\begin{aligned}
\min_{\{f_{t,p}\}}\quad
&\sum_{t\in\mathrm I_c}
 \Bigl\|\sum\nolimits_{p\in\mathrm P}f_{t,p}
 -F_t^{\mathrm{req}}\Bigr\|_2^2\\
&+\ell_\tau^{-2}\sum_{t\in\mathrm I_c}
 \Bigl\|\sum\nolimits_{p\in\mathrm P}r_{t,p}\times f_{t,p}
 -\tau_t^{\mathrm{req}}\Bigr\|_2^2\\
&+\lambda_f\sum_{t\in\mathrm I_c}\sum_{p\in\mathrm P}
 \|f_{t,p}\|_2^2\\
&+\lambda_{\Delta f}
 \sum_{\substack{t\in\mathrm I_c\\t>t_{\mathrm{on}}}}
 \sum_{p\in\mathrm P}\|f_{t,p}-f_{t-1,p}\|_2^2,\\
\mathrm{s.t.}\quad
&f^n_{p,\min}\le f^n_{t,p}\le f^n_{p,\max},\\
&\|f^\parallel_{t,p}\|_2\le\mu_p f^n_{t,p}.
\end{aligned}
\label{eq:force_allocation}
\end{equation}
This expands Eq.~\eqref{eq:contact_force_design}: the last two terms form $R_f$ and the constraints define $\mathrm F$. The norm is $\|u\|_D^2=u^{\mathsf T}Du$, with $D=\operatorname{diag}(I_3,\ell_\tau^{-2}I_3)$ converting moment error to a force scale. Weights $\lambda_f,\lambda_{\Delta f}$ penalize force magnitude and variation; $f^n_{p,\min},f^n_{p,\max}$ bound compression, and $\mu_p$ is friction.

\subsection{Anchor Adjustment}
\label{app:anchor_adjustment}
Let $U\in\mathbb R^{3\times2}$ be a shared orthonormal tangent basis and $d$ the unit plane normal of the opposing regions. Parameterize $a_p(\xi_p)=U\xi_p+b_pd$, where $b_p=d^{\mathsf T}a_p^{(0)}$ is fixed and $\xi_p^{(0)}=U^{\mathsf T}a_p^{(0)}$. With forces $\widehat f_{t,p}$ fixed and $r_{t,p}(\xi_p)=R_{O,t}^Ha_p(\xi_p)$, solve
\begin{equation}
\begin{aligned}
\min_{\{\xi_p\}}\quad
&\ell_\tau^{-2}\sum_{t\in\mathrm J}
\Bigl\|\sum\nolimits_{p\in\mathrm P}r_{t,p}(\xi_p)
\times\widehat f_{t,p}-\tau_t^{\mathrm{req}}\Bigr\|_2^2\\
&+\ell_a^{-2}\sum_{p\in\mathrm P}
\bigl\|\xi_p-\xi_p^{(0)}\bigr\|_2^2,\\
\mathrm{s.t.}\quad
&\xi_p^{\min}\le\xi_p\le\xi_p^{\max}.
\end{aligned}
\label{eq:anchor_refinement}
\end{equation}
The set $\mathrm J$ samples every other frame from $t_{\mathrm{on}}$ in $\mathrm I_c$. Scale $\ell_a$ controls displacement; componentwise bounds $\xi_p^{\min},\xi_p^{\max}$ follow the region's tangent-plane extent. Each part shares one anchor across sampled frames. Updated anchors are projected onto their regions with $n_p$ fixed. Three allocation--adjustment--projection rounds precede a final force allocation.

\subsection{Contact Coordinates and Linearization}
\label{app:contact_linearization}
Let $(R_p(q),z_p(q))$ and $(R_{O,t},z_{O,t})$ denote the robot-part and prescribed object poses in the target world frame. The contact position and direction in object coordinates are
\begin{equation}
\begin{aligned}
h_{t,p}(q)&=R_{O,t}^{\mathsf T}
 [z_p(q)+R_p(q)c_p-z_{O,t}],\\
v_{t,p}(q)&=R_{O,t}^{\mathsf T}R_p(q)u_p.
\end{aligned}
\label{eq:contact_coordinates}
\end{equation}
The fixed targets are $a_p^\star=a_p-d_{0,p}n_p$ and $-n_p$, with inward offset $d_{0,p}\ge0$. Let $J_p^W(s)$ be the world-position Jacobian of a point with local coordinates $s$. At $q_t^{(k)}$, the contact Jacobians are
\begin{equation}
\begin{aligned}
J_{h,t,p}&=R_{O,t}^{\mathsf T}J_p^W(c_p),\\
J_{v,t,p}&=R_{O,t}^{\mathsf T}
 [J_p^W(u_p)-J_p^W(0)].
\end{aligned}
\label{eq:contact_jacobians}
\end{equation}
The difference in $J_{v,t,p}$ removes translation. These Jacobians yield Eq.~\eqref{eq:contact_objectives}, with activation from Eq.~\eqref{eq:contact_activation} independent of $\gamma_e(t)$. SQP updates the robot configuration; offline refinement determines the fixed targets.

\subsection{Interaction Linearization and Regularization}
\label{app:interaction_linearization}
Source and robot vertices $X_t^H,X_t^R(q)\in\mathbb R^{V\times3}$ include object samples in object coordinates and share the source graph's fixed connectivity and uniform Laplacian $L_t$. With $\delta_t^H=L_tX_t^H$ and $\delta_t^R(q)=L_tX_t^R(q)$, Eq.~\eqref{eq:retargeting} uses
\begin{equation}
r_{t,i}^{(k)}(\Delta q)=
\delta_{t,i}^{R,(k)}+
J_{\delta,t,i}^{(k)}\Delta q-\delta_{t,i}^H,
\label{eq:interaction_linearization}
\end{equation}
where $J_{\delta,t,i}^{(k)}$ is the Jacobian of $\delta_{t,i}^{R,(k)}$. Stacking gives $J_\delta=(L_t\otimes I_3)J_X$, with Kronecker product $\otimes$ and stacked vertex-position Jacobians $J_X$; prescribed object points have zero rows. The main-text objective omits the iteration superscript.

Table~\ref{tab:str_parameters} lists regularization and constraint parameters. The solver retains linearized non-penetration, stance-foot XY constraints, configuration limits, and the increment bound. Failed initialization is retried without the increment bound. Accepted updates normalize the base quaternion and relinearize; solving stops at convergence or the frame's budget.

\subsection{Contact-Target Transfer to Policy Learning}
\label{app:rl_contact_details}
Object-local position targets and activation curves are resampled from the STR contact artifact to the RL reference clock. The G1 training model uses half-sphere hand collision geometry. Its support points are expressed in the parent wrist frames after fixed-link merging. This adaptation changes the robot-side contact points without reoptimizing the reference poses.

\Needspace*{9\baselineskip}
Let $\mathcal A_n=\{p\in\mathrm P:\rho_{t(n),p}>0\}$ denote the active contact set. For $\mathcal A_n\ne\varnothing$, the error and activation in Eq.~\eqref{eq:contact_reward} are
\begin{equation}
\begin{aligned}
e_{c,n}^2
&=\frac{\sum_{p\in\mathcal A_n}\rho_{t(n),p}
\|p_{n,p}-z_{n,p}\|_2^2}
{\sum_{p\in\mathcal A_n}\rho_{t(n),p}},\\
\gamma_n
&=\frac{1}{|\mathcal A_n|}\sum_{p\in\mathcal A_n}\rho_{t(n),p}.
\end{aligned}
\label{eq:rl_contact_aggregation}
\end{equation}
For $\mathcal A_n=\varnothing$, set $e_{c,n}=\gamma_n=0$. Binary activation gives the mean squared error over active parts and $\gamma_n=1$; fractional activation weights the errors and scales transition rewards. The exponential is applied after aggregation. The total reward in Eq.~\eqref{eq:rl_total_reward} is multiplied by the control-step duration.

\section{Real-World Deployment Configuration}
\label{app:deployment_configuration}

The Unitree R1 runs four PPO actors for pickup, rising, walking, and placement on a Jetson Orin NX. Trained in IsaacSim, the actors share a reference trajectory and are exported to ONNX with learned observation normalization. Table~\ref{tab:deployment_parameters} lists the verified deployment settings.


\begin{table}[H]
\centering
\caption{R1 policy and deployment settings. Rates are configured values.}
\label{tab:deployment_parameters}
\begingroup\footnotesize
\setlength{\tabcolsep}{3pt}\renewcommand{\arraystretch}{1.00}
\begin{tabularx}{\linewidth}{@{}>{\raggedright\arraybackslash}p{0.21\linewidth}>{\raggedright\arraybackslash}X@{}}
\toprule
\textbf{Item} & \textbf{Configuration}\\
\midrule
Robot & R1: 26 joints (legs 12, arms 10, waist 2, head 2)\\
Compute & Jetson Orin NX, 16\,GB\\
Software & Ubuntu 20.04.5; L4T 35.3.1; Linux 5.10.104-tegra; Python 3.10.21; ONNX Runtime 1.23.2 (CPU)\\
\midrule
Training & PPO / IsaacSim; 4,096 environments; 200\,Hz physics, 50\,Hz control\\
Actor & 139 inputs, 26 outputs; MLP 512--256--128, ELU; four stage-specific ONNX models\\
Observations & Previous actions (26), base angular velocity (3), joint positions/velocities (26 each), reference joint positions/velocities (52), relative reference orientation (6)\\
Normalization & Learned empirical normalization in ONNX; one-frame history\\
\midrule
Control / link & 50\,Hz policy; 500\,Hz commands; Unitree SDK2 DDS via a C++ realtime bridge\\
Joint mapping & DDS slots 0--13, 15--19, 22--26, 29, 30\\
Action scale & $0.25\times$ torque limit / $K_p$; 0.15--0.825\,rad per action unit\\
PD gains $(K_p,K_d)$ & Hips/knees/waist: $(100,2)$; ankles and shoulder pitch/roll: $(40,2)$; shoulder yaw/elbows/wrist roll: $(20,1)$; head pitch: $(50,2)$; head yaw: $(10,0.1)$\\
Limits & Joint-specific position bounds; torque clipping at 60\,Nm (hips/knees/waist/shoulder pitch/roll), 50\,Nm (ankles), 33\,Nm (others)\\
\midrule
Measurements & Joint encoders; IMU attitude/angular velocity; no measured object pose or base translation\\
Reference clock & Robot-source tick, 20\,ms spacing; query reference before observation assembly; stage gates at frames 144, 184, 304, 424\\
\bottomrule
\end{tabularx}
\endgroup
\end{table}

\end{document}